\documentclass[11pt,a4paper]{article}
\usepackage[hyperref]{emnlp2020}
\usepackage{times}
\usepackage{latexsym}

\usepackage{amsmath}
\usepackage{multirow}
\usepackage{url}
\usepackage{graphicx}
\usepackage{amssymb}
\usepackage{amsfonts}
\usepackage[noend]{algpseudocode}
\usepackage{algorithmicx,algorithm}

\usepackage{microtype}

\aclfinalcopy 

\title{Life-long Learning for Multilingual Neural Machine Translation with Knowledge Distillation}

\author{
Yang Zhao$^{1,2}$, 
Junnan Zhu$^{1,2}$, 
Lu Xiang$^{1,2}$, 
Jiajun Zhang$^{1,2}$,\AND
Yu Zhou$^{1,3}$,
Feifei Zhai$^{1,3}$,
and Chengqing Zong$^{1,2}$
\\ 
$^1$National Laboratory of Pattern Recognition, Institute of Automation, CAS, Beijing, China\\
$^2$University of Chinese Academy of Sciences, Beijing, China\\
$^3$Fanyu AI Research, Beijing Fanyu Technology Ltd., Beijing, China\\
\{yang.zhao, junnan.zhu, lu.xiang, jjzhang, yzhou, cqzong\} @nlpr.ia.ac.cn \\ feifeizhai@zkyf.com
}

\date{}

\begin{document}
\maketitle
\begin{abstract}
A common scenario of Multilingual Neural Machine Translation (MNMT) is that each translation task arrives in a sequential manner, and the training data of previous tasks is unavailable. In this scenario, the current methods suffer heavily from catastrophic forgetting (CF). To alleviate the CF, we investigate knowledge distillation based life-long learning methods. Specifically, in one-to-many scenario, we propose a multilingual distillation method to make the new model (student) jointly learn multilingual output from old model (teacher) and new task. In many-to-one scenario, we find that direct distillation faces the extreme partial distillation problem, and we propose two different methods to address it: pseudo input distillation and reverse teacher distillation. The experimental results on twelve translation tasks show that the proposed methods can better consolidate the previous knowledge and sharply alleviate the CF. 

\end{abstract}

\section{Introduction}

Recently, multilingual neural machine translation (MNMT) \cite{dong2015multi,firat2016multi,johnson2017google,gu2018universal,aharoni2019massively} draws much attention due to its remarkable improvement on low-resource language pairs and simple implementation, especially the approach with one universal encoder and decoder \cite{johnson2017google,aharoni2019massively}.

The current MNMT methods are studied in the conventional setting that bilingual pairs for all the translation tasks are available at training time. In practice, however, we always face an incremental scenario where each task arrives in a sequential manner. Assuming that we have already built an MNMT model (old model) from English to Italy and Dutch (\texttt{EN$\Rightarrow$IT, NL}), and we hope to extend the model with English-to-Romanian (\texttt{EN$\Rightarrow$RO}) translation. Generally, two basic methods can be adopted: 

i) \textbf{Fine-tuning.} We can fine-tune the old system with the new translation data.This method suffers from severe degradation on previous translation tasks, this phenomenon is known as \textbf{Catastrophic Forgetting (CF)} \cite{MCCLOSKEY1989109}. Table 1 shows the results, where after fine-tuning with \texttt{EN$\Rightarrow$RO} task, BLEU scores of the previous tasks sharply drop from 28.51 to 1.08 (\texttt{EN$\Rightarrow$IT}) and from 29.79 to 0.99 (\texttt{EN$\Rightarrow$NL}), respectively.

ii) \textbf{Joint training.} We can train a new task jointly with the previous and new training data. This method can achieve good performance. While as the number of translation tasks grows, storing and retraining on all training data becomes infeasible and cumbersome \cite{li2017learning}. More seriously, in many cases, the training data for previously learned tasks is unavailable due to the data privacy and protection \cite{shokri2015privacy,Yang2019Federated}, and it is impossible to jointly train an MNMT model under this situation.   

\begin{table}
	\small
	\centering
	\begin{tabular}{cccc}
		\hline
		Model & \texttt{EN$\Rightarrow$IT} & \texttt{EN$\Rightarrow$NL} & \texttt{EN$\Rightarrow$RO}  \\
		\hline
		Initial & 28.11 & 29.79 & $\sim$  \\
		Fine-tuning & 1.08 & 0.99 & 25.96   \\
		Joint training & 30.46 & 31.48 & 27.55 \\ 
		\hline
	\end{tabular}
    \caption{The BLEU scores of fine-tuning and joint training method. Fine-tuning method suffers from CF.}
\end{table}

Life-long learning aims at adapting a learned model to a new task while retaining the previous knowledge without accessing the previous training data. Among them, knowledge distillation based methods \cite{li2017learning,hou2018lifelong,belouadah2019il2m} are very common ways. In these methods when a new task arrives, the new model (student) is jointly learned by the old model's output (teacher) and new task. However, these methods are specifically designed for image classification \cite{li2017learning,hou2018lifelong,aharoni2019massively}, or object detection \cite{Shmelkov_2017_ICCV}, the incremental MNMT scenarios are not studied.


Therefore, in this paper, we focus on the life-long learning for incremental MNMT, and the scenarios we study here are setting as follows:

i) Each task arrives in a sequential manner.
	
ii) Training data of previous tasks is unavailable. We can only access the training data of new translation task and a learned old MNMT model.
	
iii) A single MNMT model needs to perform well on all tasks after learning a new one.

Specifically, two common incremental MNMT scenarios are considered:

\textbf{Incremental one-to-many scenario:} An MNMT model incrementally learns to translate one same source language into different target languages.

\textbf{Incremental many-to-one scenario:} An MNMT model incrementally learns to translate different source languages into one same target language\footnote{Incremental many-to-many scenario is left for future work.}.

In incremental one-to-many scenario, we propose a \textbf{multilingual distillation} method, in which the old model is treated as a teacher and the new model is treated as a student. To distillate the multilingual knowledge in the teacher, we first add the corresponding indicator of learned languages in the beginning of source sentence, which then be fed into the teacher model to get the multilingual outputs. Finally, the student model is jointly learned by the new task and multilingual distillation results.

In incremental many-to-one scenario, we find that direct distillation faces the \textbf{Extreme Partial Distillation} problem:

\textbf{Extreme Partial Distillation}: Given an old many-to-one model (such as \texttt{IT, NL$\Rightarrow$ EN}), and a new task (such as \texttt{RO$\Rightarrow$EN}), if we treat the old model as a teacher and directly input the new source sentences (\texttt{RO}) into it, the teacher model actually is fed by a sentence filled with \texttt{UNK}s due to the \texttt{UNK} replacing strategy\footnote{Most tokens in new source language (\texttt{RO}) are not appeared before, and these tokens are replaced by a special token \texttt{UNK}.}. Ideally, we hope that the student model could learn the knowledge from teacher on various tokens (whole knowledge) of previous source languages (\texttt{IT,NL}).
While in this situation, the student model can only learn from teacher how to handle \texttt{UNK}s (partial knowledge). Therefore, we define this problem as extreme partial distillation.

To address this problem in many-to-one scenario, we propose two methods: 1) \textbf{pseudo input distillation}, and 2) \textbf{reverse teacher distillation}. In the former one, we still utilize the old many-to-one MNMT model as a teacher. While instead of directly inputting the new source sentences (\texttt{RO}) into it, we first construct pseudo inputs by replacing the new tokens (\texttt{RO}) with learned tokens (\texttt{IT,NL}) via a frequency mapping. Then pseudo inputs are utilized to distillate the knowledge from teacher. In the later one, we utilize the reversed one-to-many model (\texttt{EN $\Rightarrow$ IT,NL}) as a teacher. When a new task arrives (\texttt{RO$\Rightarrow$EN}), we input the target language (\texttt{EN}) into the teacher and get the multilingual source outputs (\texttt{IT,NL}). We test the proposed methods on twelve different translation tasks. The experimental results show that the proposed methods can sharply alleviate the CF.

The contributions of this paper are listed as follows:

i) We focus on the incremental MNMT scenario and investigate the knowledge distillation based life-long learning method.

ii) In one-to-many scenario, we propose a multilingual distillation method to make the new model jointly learn multilingual output from old model and new task. 

iii) In many-to-one scenario, we find that direct distillation faces the extreme partial distillation problem, and propose two different methods (pseudo input distillation and reverse teacher distillation) to address it.

\begin{figure*}[!t]
	\centering
	\includegraphics[width=1.8\columnwidth]{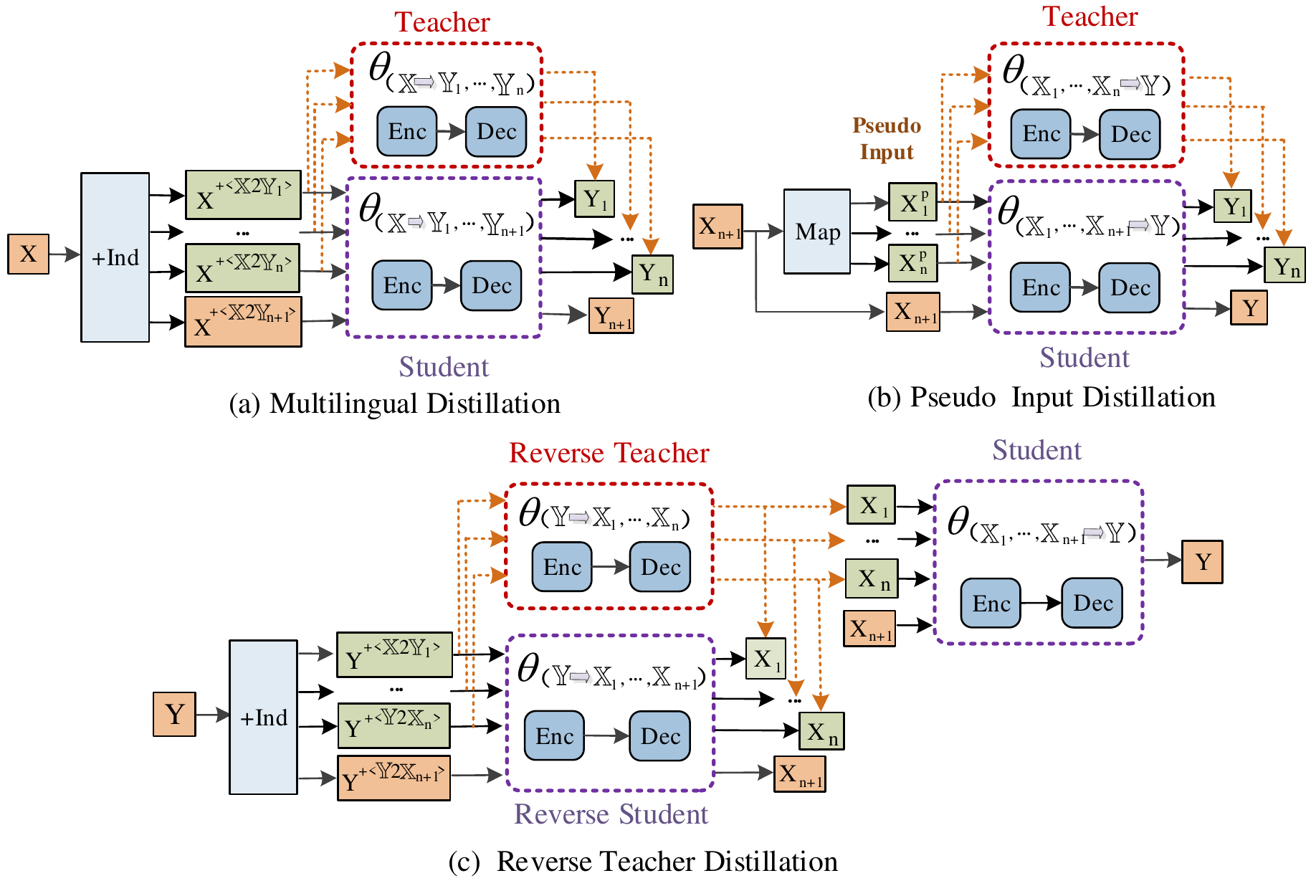}
	\caption{The framework of proposed methods, where multilingual distillation (a) is proposed for  incremental one-to-many scenario. Pseudo input distillation (b) and reverse teacher distillation (c) are proposed for incremental many-to-one scenario.}
	\label{overview}
\end{figure*}

\section{Multilingual NMT}

To make full use of multilingual data within a single system, various MNMT methods are proposed \cite{firat2016multi,johnson2017google,gu2018universal}, where \cite{johnson2017google} propose a simple while effective MNMT method. In this method, it is no need to change the network architecture. The only modification is that they introduce a special indicator at the beginning of the source sentence to indicate source and target language.

For example, consider the following English-to-Italy sentence pair:

\emph{you probably saw it on the news . $\rightarrow$  forse lo avete visto sui notiziari .}

It will be modified to:

\emph{\texttt{<en2it>} you probably saw it on the news . $\rightarrow$  forse lo avete visto sui notiziari .}
where \texttt{<en2it>} is an indicator to show that the source is English and the target is Italy.

\textbf{Notation:} We denote a one-to-many MNMT model by $\theta_{(\mathbb{X} \Rightarrow \mathbb{Y}_{1},...,\mathbb{Y}_{n})}$, where $\mathbb{X}$ is a source language, and $\mathbb{Y}_{1},...,\mathbb{Y}_{n}$ denotes $n$ different target languages. Similarly, 
a many-to-one MNMT model is denoted by $\theta_{(\mathbb{X}_{1},...,\mathbb{X}_{n} \Rightarrow \mathbb{Y})}$. We denote a translation task from $\mathbb{X}$ to $\mathbb{Y}$ by $\mathbb{X} \Rightarrow \mathbb{Y}$, whose training sentence pairs are denoted by $D_{\mathbb{X} \Rightarrow \mathbb{Y}}=\left \{ (X,Y) \right \}$, where $X$ is the source sentence and $Y$ is the target sentence. When we adding a indicator \texttt{<}$\mathbb{X}2\mathbb{Y}_{j}$\texttt{>} for target language $\mathbb{Y}_{j}$ into a source sentence $X$, we denote the source sentence by $X^{+\texttt{<}\mathbb{X}2\mathbb{Y}_{j}\texttt{>}}$.

\section{Method Description}

\subsection{Incremental One-to-many Scenario}
In incremental one-to-many scenario, given an old one-to-many model $\theta_{(\mathbb{X} \Rightarrow \mathbb{Y}_{1},...,\mathbb{Y}_{n})}$
and the training data $D_{\mathbb{X} \Rightarrow \mathbb{Y}_{n+1}}=\left \{ (X,Y_{n+1}) \right \}$ of a new task $\mathbb{X} \Rightarrow \mathbb{Y}_{n+1}$, our goal is to get a new model $\theta_{(\mathbb{X} \Rightarrow \mathbb{Y}_{1},...,\mathbb{Y}_{n+1})}$. To achieve this, we propose a multilingual distillation method to let the new model (student) jointly learn the new task and multilingual knowledge in the old model (teacher). Fig. 1 (a) illustrates the framework, which contains three steps:

\textbf{Step 1:} For each learned target language $\mathbb{Y}_{i} \in (\mathbb{Y}_{1},...,\mathbb{Y}_{n})$, we first add the indicator $\texttt{<}\mathbb{X}2\mathbb{Y}_{i}\texttt{>}$ into source sentence $X$. We denote the source sentence with indicator by $X^{+ \texttt{<}\mathbb{X}2\mathbb{Y}_{i}\texttt{>}}$.

\textbf{Step 2:} For each $X^{+ \texttt{<}\mathbb{X}2\mathbb{Y}_{i}\texttt{>}}, i \in [1,n]$, we input it into the old model $\theta_{(\mathbb{X} \Rightarrow \mathbb{Y}_{1},...,\mathbb{Y}_{n})}$ (\textbf{teacher})
and get the corresponding result $Y_{i}$ with beam search:
\begin{equation}
\small
Y_{i} \leftarrow \text{BeamSearch}(X^{+ \texttt{<}\mathbb{X}2\mathbb{Y}_{i}\texttt{>}}, \theta_{(\mathbb{X} \Rightarrow \mathbb{Y}_{1},...,\mathbb{Y}_{n})})
\end{equation}

\textbf{Step 3:} Train a new model $\theta_{(\mathbb{X} \Rightarrow \mathbb{Y}_{1},...,\mathbb{Y}_{n+1})}$ (\textbf{student}) by maximizing the following objective function: 
\begin{equation}
\small
\begin{aligned}
& L (\theta_{(\mathbb{X}  \Rightarrow  \mathbb{Y}_{1},...,\mathbb{Y}_{n+1})})  = \underbrace{ log \ p(Y_{1}|X^{+ \texttt{<}\mathbb{X}2\mathbb{Y}_{1}\texttt{>}})}_{\texttt{task} \ \mathbb{X} \Rightarrow \mathbb{Y}_{1}} + ... \\ 
	& + \underbrace{ log \ p(Y_{n}|X^{+ \texttt{<}\mathbb{X}2\mathbb{Y}_{n}\texttt{>}})} _{\texttt{task} \ \mathbb{X} \Rightarrow \mathbb{Y}_{n}} 
	+ \underbrace{ log \ p(Y_{n+1}|X^{+ \texttt{<}\mathbb{X}2\mathbb{Y}_{n+1}\texttt{>}})}_{\texttt{new task} \ \mathbb{X} \Rightarrow \mathbb{Y}_{n+1}}
\end{aligned}
\end{equation}
where the first $n$ items denote the loss of previous tasks produced by the teacher. The last one denotes the loss of new task. 

\textbf{Discussion.} We call this method as multilingual distillation, since the new model could learn multilingual knowledge in the old model. During distillation, we can also utilize the $k$-best sentence distillation ($k>1$) \cite{kim2016sequence,tan2019multilingual} or greedy search distillation. The experimental results can be found in Sec. 5.1.

\subsection{Incremental Many-to-one Scenario} 
In this scenario, our goal is to get a new model $\theta_{(\mathbb{X}_{1},...,\mathbb{X}_{n+1} \Rightarrow \mathbb{Y})}$ with an old many-to-one model $\theta_{(\mathbb{X}_{1},...,\mathbb{X}_{n} \Rightarrow \mathbb{Y})}$
and training sentence pairs $D_{\mathbb{X}_{n+1} \Rightarrow \mathbb{Y}}=\left \{ (X_{n+1},Y) \right \}$ of a new task $\mathbb{X}_{n+1} \Rightarrow \mathbb{Y}$. As we mentioned before, direct distillation faces the extreme partial distillation problem. To address this, we propose two different methods: 1) pseudo input distillation  (Fig. 1 (b)), and 2) reverse teacher distillation (Fig. 1 (c)).  

\subsubsection{Pseudo Input Distillation} 
In this method, instead of inputting $X_{n+1}$ into the teacher, we first construct a pseudo input by transforming the new token of $X_{n+1}$ into the previous tokens. Then the pseudo input is utilized to distillate the knowledge. Specifically, this method contains four steps:

\textbf{Step 1:} For each source language $\mathbb{X}_{i} \in (\mathbb{X}_{1},...,\mathbb{X}_{n+1})$, we sort its vocabulary $V_{\mathbb{X}_{i}}$ in descending order by the frequency as follows:
\begin{equation}
\small
\begin{aligned}
V_{\mathbb{X}_{i}}=&\left \{s^{1}_{\mathbb{X}_{i}},..., s^{j}_{\mathbb{X}_{i}}, ..., s^{|V_{\mathbb{X}_{i}}|}_{\mathbb{X}_{i}}\right \} \\
\text{where} \ \ N(s^{1}_{\mathbb{X}_{i}})> \ & ... \ >N(s^{j}_{\mathbb{X}_{i}})> \ ... \ >N(s^{|V_{\mathbb{X}_{i}}|}_{\mathbb{X}_{i}})
\end{aligned}
\end{equation}
$N(s^{j}_{\mathbb{X}_{i}})$ represents the frequency of token $s^{j}_{\mathbb{X}_{i}}$ in the corresponding training data, and $j$ is the frequency ranking of this token in all vocabulary.

Then we can construct a mapping between a new token $s^{j}_{\mathbb{X}_{n+1}}$ and a learned token $s^{j}_{\mathbb{X}_{i}}$, if these two tokens have the same ranking $j$. Formally, 
\begin{equation}
\small
M_{(\mathbb{X}_{n+1} \rightarrow \mathbb{X}_{i})}: s^{j}_{\mathbb{X}_{n+1}} \rightarrow s^{j}_{\mathbb{X}_{i}}, \forall j,i
\end{equation}
where $s^{j}_{\mathbb{X}_{n+1}}$ is a new token in language $\mathbb{X}_{n+1}$ which ranked $j$th in $V_{\mathbb{X}_{n+1}}$ and $s^{j}_{\mathbb{X}_{i}}$ is a learned token in language $\mathbb{X}_{i}$ which also ranked $j$th in $V_{\mathbb{X}_{i}}$

\textbf{Step 2:} Given a new source sentence $X_{n+1}$, we replace the tokens in $X_{n+1}$ with learned tokens in languages $\mathbb{X}_{i}$  with the mapping $M_{(\mathbb{X}_{n+1} \rightarrow \mathbb{X}_{i})}$ (Eq. (4)) by

\begin{equation}
\small
X^{p}_{i} \leftarrow \text{Mapping}(X_{n+1}, M_{(\mathbb{X}_{n+1} \rightarrow \mathbb{X}_{i})}),  i\in [1,n] 
\end{equation}
where $X^{p}_{i}$ is the pseudo input, which contains the tokens of language $\mathbb{X}_{i}$.

\textbf{Step 3:} For each pseudo input $X^{p}_{i} \in (X^{p}_{1},...,X^{p}_{n}) $, we input it into the old model  $\theta_{(\mathbb{X}_{1},...,\mathbb{X}_{n} \Rightarrow \mathbb{Y})}$ (teacher)
and get the distillation result $Y_{i}$ with beam search:
\begin{equation}
\small
Y_{i} \leftarrow \text{BeamSearch}(X^{p}_{i}, \theta_{(\mathbb{X}_{1},...,\mathbb{X}_{n} \Rightarrow \mathbb{Y})})
\end{equation}

\textbf{Step 4:} Train a new model $\theta_{(\mathbb{X}_{1},...,\mathbb{X}_{n+1} \Rightarrow \mathbb{Y})}$ (student) by maximizing the following objective function: 
\begin{equation}
\small
\begin{aligned}
L( & \theta_{( \mathbb{X}_{1},...,\mathbb{X}_{n+1} \Rightarrow \mathbb{Y})})  = \underbrace{ log \ p(Y_{1}|X^{p}_{1})  }_{\texttt{task} \ \mathbb{X}_{1} \Rightarrow \mathbb{Y}} + ...\\
& + \underbrace{ log \ p(Y_{n}|X^{p}_{n})}_{\texttt{task} \ \mathbb{X}_{n} \Rightarrow \mathbb{Y}} + \underbrace{ log \ p(Y_{n+1}|X_{n+1})}_{\texttt{new task} \ \mathbb{X}_{n+1} \Rightarrow \mathbb{Y}}
\end{aligned}
\end{equation}
where the first $n$ items denote the loss of previous tasks produced by the teacher. The last one denotes the loss of new task.

\textbf{Discussion.} This pseudo input distillation could alleviate extreme partial distillation problem, since the pseudo inputs $X^{p}_{i}$ contains the various tokens of previous learned language. Thus, when we utilize $X^{p}_{i}$ as the distillation input, the student model could learn from teacher that how to translate these tokens while not just \texttt{unk}s. Meanwhile, the additional cost of this method is small, we only need to maintain a sorted vocabulary in descending order for each learned source languages. 
   
\subsubsection{Reverse Teacher Distillation} 
In reverse teacher distillation, beside the old many-to-one model $\theta_{(\mathbb{X}_{1},...,\mathbb{X}_{n} \Rightarrow \mathbb{Y})}$, we also need a reverse one-to-many model $\theta_{(\mathbb{Y} \Rightarrow \mathbb{X}_{1},...,\mathbb{X}_{n})}$ at the same time. To alleviate the extreme partial distillation problem, when the training data of a new task $D_{\mathbb{X}_{n+1} \Rightarrow \mathbb{Y}}=\left \{ (X_{n+1},Y) \right \}$ arrives, we treat the one-to-many model $\theta_{(\mathbb{Y} \Rightarrow \mathbb{X}_{1},...,\mathbb{X}_{n})}$ as a teacher, and input the target sentence $Y$ in it. Specifically, this method contains four steps:

\textbf{Step 1:} For each learned source languages $\mathbb{X}_{i} \in (\mathbb{X}_{1},...,\mathbb{X}_{n})$, we first add the indicator $\texttt{<}\mathbb{Y}2\mathbb{X}_{i}\texttt{>}$ into target sentence $Y$, and denote the target sentence with indicator by $Y^{+\texttt{<}\mathbb{Y}2\mathbb{X}_{i}\texttt{>}}$.

\textbf{Step 2:} For target sentence with indicator $Y^{+\texttt{<}\mathbb{Y}2\mathbb{X}_{i}\texttt{>}} \in (Y^{+\texttt{<}\mathbb{Y}2\mathbb{X}_{1}\texttt{>}},...,Y^{+\texttt{<}\mathbb{Y}2\mathbb{X}_{n}\texttt{>}})$, we input it into the reverse one-to-many model $\theta_{(\mathbb{Y} \Rightarrow \mathbb{X}_{1},...,\mathbb{X}_{n})}$ (\textbf{reverse teacher})
and get the distillation result $X_{i}$ with beam search:
\begin{equation}
\small
X_{i} \leftarrow \text{BeamSearch}(Y^{+\texttt{<}\mathbb{Y}2\mathbb{X}_{i}\texttt{>}}, \theta_{(\mathbb{Y} \Rightarrow \mathbb{X}_{1},...,\mathbb{X}_{n})})
\end{equation}

\textbf{Step 3:} Train a new many-to-one model (\textbf{student})  $\theta_{(\mathbb{X}_{1},..., \mathbb{X}_{n+1} \Rightarrow \mathbb{Y})}$ by maximizing the following objective function: 
\begin{equation}
\small
\begin{aligned}
L(& \theta_{(\mathbb{X}_{1},..., \mathbb{X}_{n+1} \Rightarrow \mathbb{Y})})=
 \underbrace{ log \ p(Y|X_{1})}_{\texttt{task} \ \mathbb{X}_{1} \Rightarrow \mathbb{Y} }+ \ ...\   \\
& + \underbrace{ log \ p(Y|X_{n}) }_{\texttt{ task } \ \mathbb{X}_{n} \Rightarrow \mathbb{Y} } 
 + \underbrace{ log \ p(Y|X_{n+1})}_{\texttt{new task} \ \mathbb{X}_{n+1} \Rightarrow \mathbb{Y}}
\end{aligned}
\end{equation}
where the first $n$ items denote the loss of previous tasks produced by the reverse teacher. The last one denotes the loss of new task.

\textbf{Step 4:} Train a new reverse one-to-many model (\textbf{reverse student})  $\theta_{(\mathbb{Y} \Rightarrow \mathbb{X}_{1},..., \mathbb{X}_{n+1})}$ by maximizing the following objective function: 
\begin{equation}
\small
\begin{aligned}
L(&\theta_{(\mathbb{Y} \Rightarrow \mathbb{X}_{1},..., \mathbb{X}_{n+1})})=
 \underbrace{ log \ p(X_{1}|Y) \    }_{\texttt{task} \ \mathbb{Y} \Rightarrow \mathbb{X}_{1}} + ... \\
& + \underbrace{ log \ p(X_{n}|Y)}_{\texttt{task} \ \mathbb{Y} \Rightarrow \mathbb{X}_{n}} + \underbrace{ log \ p(X_{n+1}|Y)}_{\texttt{new task} \ \mathbb{Y} \Rightarrow \mathbb{X}_{n+1}}
\end{aligned}
\end{equation}
We also need to update the reverse student, since it will be utilized as a reverse teacher when the next task $\mathbb{X}_{n+2} \Rightarrow \mathbb{Y}$ arrives.

\textbf{Discussion.} This reverse teacher distillation could alleviate extreme partial distillation problem, since we utilize the reverse one-to-many model as a teacher and input the target into it. This idea is partially inspired by the back-translation \cite{sennrich2016improving} and generative replay methods in life-long learning \cite{shin2017continual,zhai2019lifelong}. Meanwhile, the additional cost of this method is acceptable. We need to maintain a revered one-to-many model when a new task arrives.

\section{Experimental Setting}

\textbf{Dataset.} We test the proposed methods on 12 tasks (Table 2), where English-Italian (\texttt{EN$\Leftrightarrow$IT}), English-Dutch (\texttt{EN$\Leftrightarrow$NL})  and English-Romanian (\texttt{EN$\Leftrightarrow$RO}) come from TED dataset\footnote{\url{https://wit3.fbk.eu/}}. Uygur-Chinese (\texttt{UY$\Leftrightarrow$CH}), Tibetan-Chinese (\texttt{TI$\Leftrightarrow$CH}) and Mongolian-Chinese (\texttt{MO$\Leftrightarrow$CH}) come from CCMT-19 dataset. Chinese-English (\texttt{CH$\Leftrightarrow$EN}) is LDC dataset. Japanese-English (\texttt{JA$\Leftrightarrow$EN}) is KFTT dataset\footnote{\url{http://www.phontron.com/kftt/}}. German-English (\texttt{DE$\Leftrightarrow$EN}), Finnish-English (\texttt{FI$\Leftrightarrow$EN}), Latvian-English (\texttt{LV$\Leftrightarrow$EN}) and Turkish-English (\texttt{TR$\Leftrightarrow$EN}) come from WMT-17 dataset\footnote{\url{http://data.statmt.org/wmt17/translation-task/preprocessed/}}.

\textbf{Training and Evaluation Details.} We implement our approach based on the THUMT toolkit \cite{zhang2017thumt}\footnote{\url{https://github.com/THUNLP-MT/THUMT}}. We use the ``base'' parameters in Transformer \cite{vaswani2017attention}. We use the BPE \cite{sennrich2015neural} method to merge 30K steps. For evaluation, we use beam search with a beam size of $k = 4$ and length penalty. We evaluate the translation quality with BLEU \cite{papineni2002bleu} for all tasks. 
For each task, we set both source and target vocabularies by 30K. When learning a new task, the new vocabularies are the union of previous vocabularies and vocabularies of new arrival task, i.e., $V_{(\mathbb{X}_{1},...,\mathbb{X}_{n} \Rightarrow \mathbb{Y})}=V_{(\mathbb{X}_{1},...,\mathbb{X}_{n-1} \Rightarrow \mathbb{Y})}\cup V_{(\mathbb{X}_{n} \Rightarrow \mathbb{Y})}$.

\begin{table}
	\small
	\centering
	\begin{tabular}{cccccccc}
		\hline
		Dataset & Task & Train & Dev & Test \\
		\hline 
		\hline
		\multirow{3}{*}{TED} & \texttt{EN$\Leftrightarrow$IT} & 232k & 929 & 1566\\
		&  \texttt{EN$\Leftrightarrow$NL}  & 237k & 1003 & 1777\\
		&  \texttt{EN$\Leftrightarrow$RO}  & 221k & 914  & 1678\\
		\hline
		\multirow{3}{*}{CCMT-19} & \texttt{MO$\Leftrightarrow$CH} & 254k & 2000 & 1000 \\
		& \texttt{TI$\Leftrightarrow$CH} & 155k & 2000 & 1000 \\
		& \texttt{UY$\Leftrightarrow$CH} & 168k & 2000 & 1000 \\
		\hline
		LDC & \texttt{CH$\Leftrightarrow$EN} & 2.1M & 919 & 6146 \\
		\hline
		KFTT & \texttt{JA$\Leftrightarrow$EN} & 440k & 1166 & 1160 \\
		\hline
		\multirow{4}{*}{WMT-17} & \texttt{DE$\Leftrightarrow$EN} & 5.9M & 3003 & 5168 \\
		& \texttt{FI$\Leftrightarrow$EN} & 2.6M & 2870 & 3000 \\
		& \texttt{LV$\Leftrightarrow$EN} & 4.5M & 1000 & 1003 \\
		& \texttt{TR$\Leftrightarrow$EN} & 207K & 2870 & 3000 \\
		\hline	
	\end{tabular}
	\caption{\label{tab-data} The numbers of sentence pairs in each task. }
\end{table}

\begin{table*}
	\small
	\centering
	\begin{tabular}{clccccc}
		\hline 
		\# & Model  & BLEU-task1 & BLEU-task2 & BLEU-task3 & BLEU-avg  & $\bigtriangleup$   \\
		\hline
		\hline
		\multicolumn{7}{c}{\texttt{EN$\Rightarrow$IT $\rightarrow$ EN$\Rightarrow$NL $\rightarrow$ EN$\Rightarrow$RO}} \\
		\hline
		1  & Single                & 27.76  & 28.64    &  25.27   & 27.22 & $\sim$ \\
		2  & Joint Training        & 30.46  & 31.48    & 27.55    & 29.83 & $+$2.61  \\ 
		3  & Fine-tuning           & 1.05   & 0.82     & 25.82    & 9.23  & $-$17.99 \\
		4  & EWC                   & 10.32  & 11.17    & 22.31    & 14.60  & $-$12.62 \\
		\hline
		5 & Multi-Distill (greedy) & 29.00$^{*}$ & 28.97             & 26.49$^{*}$ & 28.15 & $+$0.93 \\
		6 & Multi-Distill (beam)   & 30.31$^{*}$ & 30.11$^{\dagger}$ & 26.86$^{*}$ & 29.09 & $+$1.87 \\
		7 & Multi-Distill (2-best) & 30.10$^{*}$ & 30.37$^{\dagger}$ & 27.18$^{*}$ & 29.22 & $+$2.00 \\
		8 & Multi-Distill (4-best) & 30.51$^{*}$ & 30.52$^{*}$       & 26.81$^{*}$ & 29.28 & $+$2.06 \\
		\hline
		\hline
		\multicolumn{7}{c}{\texttt{CH$\Rightarrow$MO $\rightarrow$ CH$\Rightarrow$TI $\rightarrow$ CH$\Rightarrow$UY}}\\
		\hline
		9  & Single           & 29.19  & 29.67   & 16.20 & 25.02 & $\sim $  \\
		10  & Joint Training   & 28.97  & 29.48   & 17.51 & 25.32 & $+$0.30  \\ 
		11  & Fine-tuning     & 1.04   & 0.93    & 16.67 & 6.21  & $-$18.81 \\
		12  & EWC             & 11.32  & 12.43   & 15.53 & 13.09 & $-$11.93 \\
		\hline
		13 & Multi-Distill (greedy) & 28.88             & 29.30 & 16.52             & 24.90 & $-$0.12 \\
		14 & Multi-Distill (beam)   & 29.11             & 29.44 & 16.79$^{\dagger}$ & 25.11 & $+$0.09 \\
		15 & Multi-Distill (2-best) & 29.34             & 29.39 & 16.90$^{*}$       & 25.21 & $+$0.19 \\
		16 & Multi-Distill (4-best) & 29.50$^{\dagger}$ & 29.51 & 16.98$^{*}$       & 25.33 & $+$0.31 \\
		\hline
	\end{tabular}
    \caption{\label{tab-ccmt} The translation results in incremental one-to-many scenario. BLEU-task1, BLEU-task2 and BLEU-task3 show the BLEU scores of the corresponding three tasks. BLEU-avg is the average BLEU scores. $\bigtriangleup$ is the improvement comparing with single method. ``${\dagger}$'' indicates that the proposed system is statistically significant better ($p<0.05$) than the single system and ``*'' indicates $p<0.01$.}
\end{table*}

\textbf{Comparing methods:} We compare the proposed models against the following systems:

i) \textbf{Single:} We train each translation task with each single model by using transformer.

ii) \textbf{Joint Training:}  We implement the joint training method \cite{johnson2017google} as an upper bound of proposed life-long learning methods.  
	 
iii) \textbf{Fine-tuning:} We fine-tune the old system with the new training data of a new task.
	
iv) \textbf{EWC:} This is the Elastic Weight Consolidation (EWC) model \cite{kirkpatrick2017overcoming}. The approach injects a penalty on the difference between the parameters for the old and the new tasks into the loss function to alleviate the CF. 

v) \textbf{Multi-Distill:} This is the proposed Multilingual Distillation in one-to-many scenario. \textbf{Multi-Distill (greedy)} and \textbf{Multi-Distill (beam)} denote that during distillation, we utilize the greedy search and beam search, respectively.

vi) \textbf{Direct-Distill:} In this method, we directly utilize the new source as distillation input.

vii) \textbf{PseudoInput-Distill:} This is the proposed Pseudo Input Distillation in many-to-one scenario.  

viii) \textbf{ReverseTeacher-Distill:} This is the proposed Reverse Teacher Distillation in many-to-one scenario.

\section{Experimental Results}

\subsection{Incremental One-to-many Scenario}

\textbf{Main results.} Table 3 shows the translation results in incremental one-to-many scenario, where line 1-8 report the results that the first task is \texttt{EN$\Rightarrow$IT}, the second one is \texttt{EN$\Rightarrow$NL}, and the last one is \texttt{EN$\Rightarrow$RO}. Line 9-16 report the results that
the first task is \texttt{CH$\Rightarrow$MO}, the second one is \texttt{CH$\Rightarrow$TI}, and the last one is \texttt{CH$\Rightarrow$UY}. 
From the results, we can reach the following conclusions: 

i) Fine-tuning method suffers heavily from CF problem. Compared with the single method, the average BLEU scores sharply dropped to 9.23 (line 3) and 6.21 (line 11), respectively.
EWC method can partially alleviate CF, while the average BLEU scores remain below the single method by 12.62 (line 4) and 11.93 (line 12) , respectively.

ii) The proposed multilingual distillation method can sharply alleviate the CF. After learning three continuous tasks, the average BLEU points are 29.09 (line 6) and 25.11 (line 14), respectively. Compared with the single method, its improvement can reach to 1.87 and 0.09 BLEU points.

iii) We also investigate the $k$-best distillation in our method. We can find that $k$-best distillation ($k=2$ and $4$) can only slightly improve the current $1$-best distillation. Meanwhile, we can also find that during distillation, the beam search distillation (line 6 and 14) can achieve better results than greedy search distillation (line 5 and 13).

\begin{table}
	\small
	\centering
	\begin{tabular}{lcc}
		\hline
		Model & BLEU-prev & BLEU-new  \\
		\hline
		\hline
		Existing       & 23.68 & $\sim$      \\
		\hline
		Single         & 23.49 & 21.14  \\
		Joint Training & 23.94 & 20.88	\\ 
		Fine-tuning    & 1.07  & 21.21  \\
		\hline
		Multi-Distill  & 23.80 & 21.19 \\
		\hline
	\end{tabular}
    \caption{The translation results of an already existing on-to-many model. The existing model is trained by \texttt{EN$\Rightarrow$IT, NL, RO, JA, DE, FI, LV, TR} tasks, and the new task is \texttt{EN$\Rightarrow$CH}. }
\end{table}

\begin{table*}
	\small
	\centering
	\begin{tabular}{clccccc} 
		\# & Model  & BLEU-task1 & BLEU-task2 & BLEU-task3 & BLEU-avg & $\bigtriangleup$  \\
		\hline
		\hline
		\multicolumn{7}{c}{\texttt{IT$\Rightarrow$EN $\rightarrow$ NL$\Rightarrow$EN $\rightarrow$ RO$\Rightarrow$EN}} \\
		\hline
		1  & Single                          & 30.15  & 32.50    &  30.86  &  31.17 & $\sim$   \\
		2  & Joint Training                  & 32.04  & 34.62	 &  33.49  &  33.38 & $+$2.21  \\ 
		3  & Fine-tuning                     & 9.77   & 14.30    &  29.51  &  17.86 & $-$13.31 \\
		4  & EWC                             & 21.29  & 24.32    &  28.07  &  24.56 & $-$6.61  \\
		5  & Direct-Distill                  & 1.12   & 1.43     &  29.43  &  10.66 & $-$20.51 \\
		\hline
		6 & PseudoInput-Distill (greedy)     & 27.01        & 29.33          &  31.86$^{*}$  &  29.40 & $-$1.77 \\
		7 & ReverseTeacher-Distill (greedy)  & 31.04$^{*}$  & 33.87$^{*}$	 &  32.75$^{*}$  &  32.55 & $+$1.38 \\
		8 & PseudoInput-Distill (beam)       & 27.93        & 30.01          &  32.27$^{*}$  &  30.07 & $-$1.10 \\
		9 & ReverseTeacher-Distill (beam)    & 31.80$^{*}$  & 34.32$^{*}$	 &  32.95$^{*}$  &  33.02 & $+$1.85 \\
		\hline
		\hline
		\multicolumn{7}{c}{\texttt{MO$\Rightarrow$CH $\rightarrow$ TI$\Rightarrow$CH $\rightarrow$ UY$\Rightarrow$CH}}\\
		\hline
		10  & Single                         & 40.60  & 32.79  & 22.62  & 32.00  & $\sim$   \\
	    11  & Joint Training                 & 42.30  & 31.89  & 24.10  & 32.76  & $+$0.76  \\ 
		12  & Fine-tuning                    & 10.59  & 11.05  & 22.49  & 14.71  & $-$17.29 \\
		13  & EWC                            & 24.98  & 21.38  & 21.33  & 22.56  & $-$9.44  \\
		14  & Direct-Distill                 & 1.09   & 2.11   & 20.98  & 8.06   & $-$23.94 \\
		\hline
		15 & PseudoInput-Distill (greedy)         & 36.42        & 28.77  & 23.17              & 29.45  & $-$2.55   \\  
		16 & ReverseTeacher-Distill (greedy)      & 39.52        & 31.89  & 23.29$^{\dagger}$  & 31.57  & $-$0.43   \\
		17 & PseudoInput-Distill (beam)           & 38.01        & 30.19  & 23.30$^{\dagger}$        & 30.50  & $-$1.50   \\  
		18 & ReverseTeacher-Distill (beam)        & 42.04$^{*}$  & 32.14  & 24.01$^{*}$        & 32.73  & $+$0.73   \\
		\hline
	\end{tabular}
    \caption{The translation results in incremental many-to-one scenario. ``${\dagger}$'' indicates that the proposed system is statistically significant better ($p<0.05$) than the single system and ``*'' indicates $p<0.01$.}
\end{table*}

\textbf{Results on an already existing MNMT model.} We also conduct an experiment on the basis of an existing MNMT model. The existing model is trained by \texttt{EN$\Rightarrow$IT, NL, RO, JA, DE, FI, LV, TR} tasks, and the new task is \texttt{EN$\Rightarrow$CH}. Table 4 lists the results, where the initial average BLEU score of learned tasks is 23.68. After learning \texttt{EN$\Rightarrow$CH}, the multilingual distillation improves the BLEU score of learned tasks to 23.80 and new task to 21.19. Compared with the joint training, it achieves comparable results (23.94 vs. 23.80 and 20.88 vs. 21.19). The results show that the proposed method can consolidate the previous knowledge when learning a new task.

\begin{figure}[!t]
	\centering
	\includegraphics[width=0.97\columnwidth]{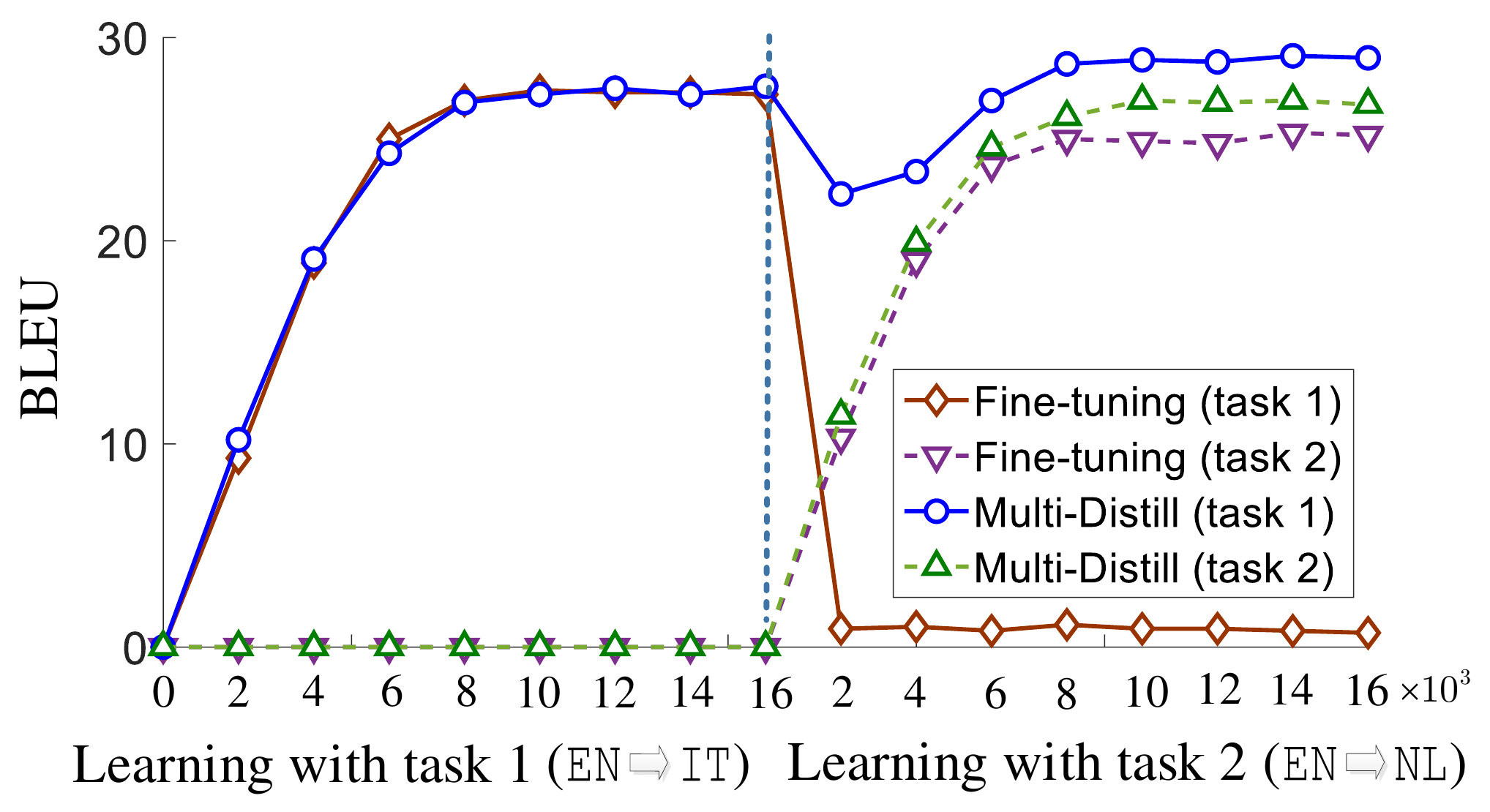}
	\caption{The BLEU scores during training in one-to-many scenario. $x$ axis denotes the training epoch and $y$ axis denotes the BLEU scores of development set.}
	\label{overview}
\end{figure}

\textbf{Results during training.} We are also curious about the results at each training epoch. Thus we record the BLEU scores of development set during training, where the first task is \texttt{EN$\Rightarrow$IT} and the second one is \texttt{EN$\Rightarrow$NL}. Fig. 2 reports the results, where $x$ axis denotes the training epoch and $y$ axis denotes the BLEU scores of development set. The results show that when we fine-tune the learned model ($\theta_{\texttt{EN$\Rightarrow$IT}}$) with \texttt{EN$\Rightarrow$NL} task, the BLEU score of the first one sharply drops to 0.99. The proposed multilingual distillation can alleviate this CF problem.

\subsection{Incremental Many-to-one Scenario}

\textbf{Main results.} Table 5 shows the translation results in incremental many-to-one scenario, where line 1-9 report the results that the first task is \texttt{IT$\Rightarrow$EN}, the second one is \texttt{NL$\Rightarrow$EN}, and the last one is \texttt{RO$\Rightarrow$EN}. Line 10-18 report the results that the first task is \texttt{MO$\Rightarrow$CH}, the second one is \texttt{TI$\Rightarrow$CH}, and the last one is \texttt{UY$\Rightarrow$CH}. From the results, we can reach the following conclusions: 

i) Fine-tuning method also suffers heavily from CF problem, whose average-BLEU scores seriously drop from 31.17 (line 1) to 17.86 (line 3) and from 32.00 (line 10) to 14.71 (line 12), respectively. We can also see that CF here is not serious as that in incremental one-to-many scenario (see Table 3). EWC method can partially alleviate CF, while it is still lower than the single model by 6.61 (line 4) and 9.44 (line 13) BLEU points, respectively. 

ii) Surprisingly, direct distillation further worsens the CF due to the extreme partial distillation, whose average-BLEU scores reduce to 10.66 (line 5) and 8.06 (line 14), respectively.

iii) The pseudo input distillation method can sharply alleviate the CF. The average BLEU scores can reach to 30.07 (line 8) and 30.50 (line 17), respectively. The reverse teacher distillation can exceed the single model by 1.85 (line 9) and 0.73 (line 18), respectively. Compared with the joint training, this method can achieve comparable results (33.02 vs. 33.38 and 32.73 vs. 32.76).

iv)  In both pseudo input distillation and reverse teacher distillation, we find that during distillation, the beam search distillation can also achieve better results than the greedy search distillation.

\begin{table}
	\small
	\centering
	\begin{tabular}{lcc}
		\hline
		Model & BLEU-prev & BLEU-new  \\
		\hline
		\hline
		Initial                &  27.34 & $\sim$ \\
		\hline
		Single                 &  27.01 & 44.40 \\
		Joint Training         &  27.41 & 44.28	\\ 
		Fine-tuning            &  1.00  & 44.53 \\
		\hline
		PseudoInput-Distill    &  23.04 & 43.87 \\
		ReverseTeacher-Distill &  27.27 & 44.25 \\
		\hline
	\end{tabular}
    \caption{The translation results of an already existing many-to-one model. The existing model is trained by \texttt{IT, NL, RO, JA, DE, FI, LV, TR$\Rightarrow$EN} tasks, and the new task is \texttt{CH$\Rightarrow$EN}.}
\end{table}

\textbf{Results on an already existing MNMT model.} We also conduct an experiment on the basis of an existing many-to-one model. Here, the existing model is trained by \texttt{IT, NL, RO, JA, DE, FI, LV, TR$\Rightarrow$EN} tasks, and the new task is \texttt{CH$\Rightarrow$EN}. Table 6 lists the results, where the initial BLEU score of learned tasks is 27.34. After learning \texttt{CH$\Rightarrow$EN}, the reverse teacher distillation can retain the BLEU score of learned tasks to 27.27. Meanwhile its BLEU score of new task is 44.25. These results are comparable with that of joint training (27.27 vs. 27.41 and 44.25 vs. 44.28). The results show that the proposed methods could also alleviate the CF in many-to-one scenario. 

\textbf{Results during training.} Fig. 3 reports the BLEU scores of development set at each training epoch, where the first task is \texttt{IT$\Rightarrow$EN} and the second one is \texttt{NL$\Rightarrow$EN}. From the results, we can see that fine-tuning also faces the CF when learning the second task. Both pseudo input distillation and reverse teacher distillation could alleviate this problem. In particular, reverse teacher distillation could further improve performance of the first task when learning the second one. Meanwhile, compared with fine-tuning method, these two methods can also improve the BLEU scores of the second task.

\begin{figure}[!t]
	\centering
	\includegraphics[width=0.97\columnwidth]{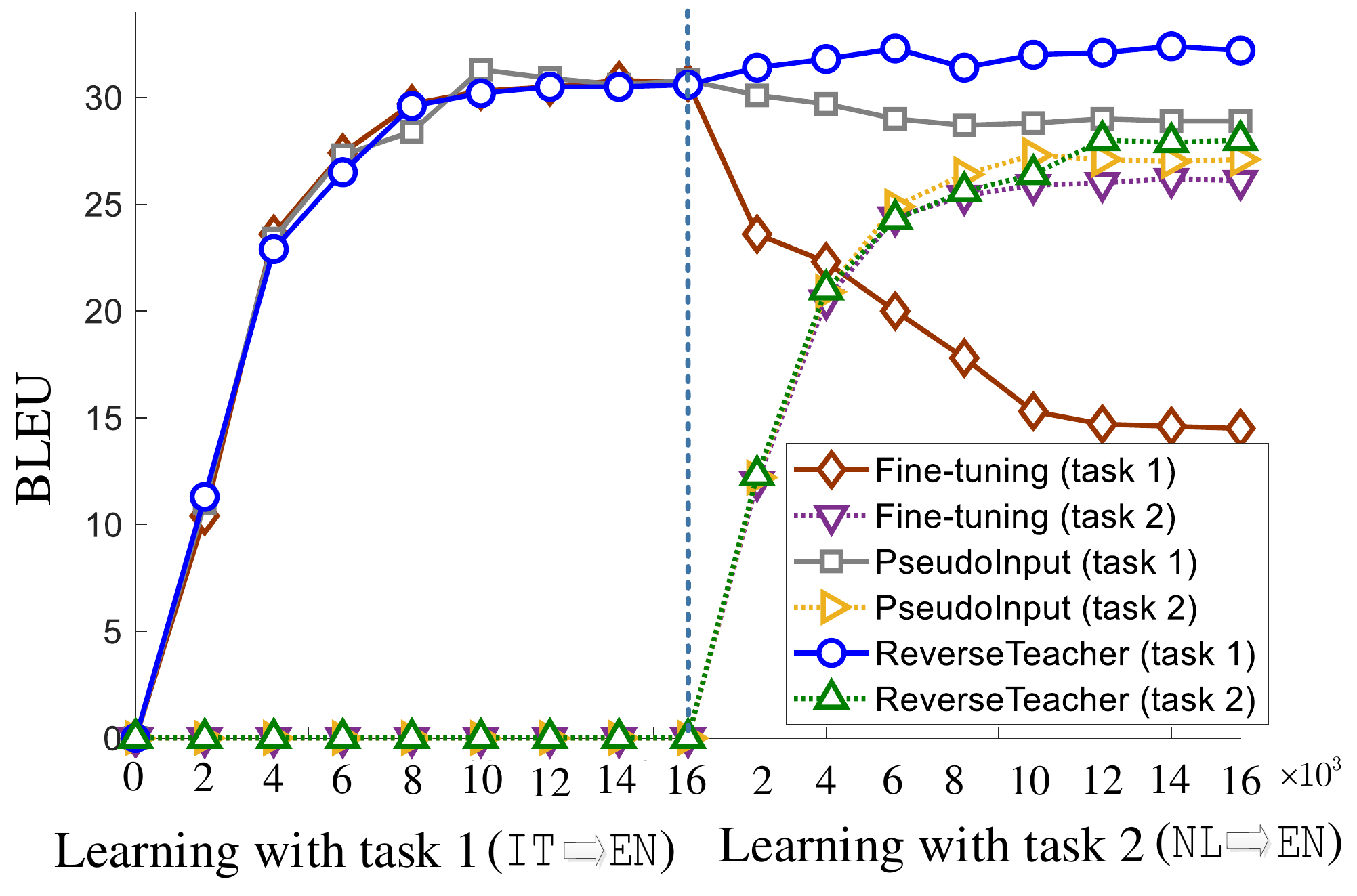}
	\caption{The BLEU scores during training in many-to-one scenario. $x$ axis denotes the training epoch and $y$ axis denotes the BLEU scores of development set.}
	\label{overview}
\end{figure}

\section{Related Work}

\textbf{Multilingual Neural Machine Translation.} To facilitate the deployment and improve the performance, various MNMT models are proposed \cite{dong2015multi,johnson2017google,gu2018universal,wang2018three,tan2019multilingual,aharoni2019massively,Yining2019A,Kudugunta2019Investigating,bapna2019simple}, where \citet{tan2019multilingual} propose a knowledge distillation method for MNMT. However, these studies are conducted in the conventional setting that bilingual pairs for all the translation tasks are available at training time. Different from these studies, we focus on incremental scenario that the training data of previous tasks is unavailable.

\textbf{Life-long Learning and Its Application in NLP.} Lifelong learning aims at adapting a learned model to new tasks while retaining the previous knowledge. \citet{de2019continual} classify these methods into three categories: i) replay-based methods \cite{lopez2017gradient,wu2018memory}, ii) regularization-based methods \cite{li2017learning}, and iii) parameter isolation-based methods \cite{rusu2016progressive}. Meanwhile, several studies apply these methods into NLP tasks, such as sentiment analysis \cite{Chen2018Lifelong,xia2017distantly}, word and sentence representation learning \cite{xu2018lifelong,liu2019continual}, language modeling \cite{sun2019lamal,de2019episodic}, domain adaptation for NMT \cite{barone2017regularization,thompson2019overcoming} and post-editors for NMT \cite{turchi2017continuous,thompson2019overcoming}. Different from these studies, we focus on the incremental MNMT scenario that each tasks arrive in a sequential manner and training data of previous tasks is unavailable. Recently, \citet{escolano2019bilingual} propose an incremental training method for MNMT, in which they train the independent encoders and decoders for each languages. While with the increasing of learning language pair, the model parameters become larger. Different from this study, we apply the life-long learning method on a more challenging MNMT framework with one universal encoder and decoder \cite{johnson2017google}.

\section{Conclusion and Future Work}
In this paper, we aim at enabling the MNMT to learn incremental translation tasks over a lifetime. To achieve this, we investigate knowledge distillation based life-long learning for MNMT. In one-to-many scenario, we propose a multilingual distillation method. In incremental many-to-one scenario, we find that direct distillation faces the extreme partial distillation problem, and propose pseudo input distillation and reverse teacher distillation to address this problem. The extensive experiments demonstrate that our method can retain the previous knowledge when learning a new task.

As a novel attempt of life-long learning for MNMT, the proposed methods still have a drawback that they cost more computational overhead due to the knowledge distillation. Therefore, in the future we will study how to reduce the computational overhead of our methods and extend them into the incremental many-to-many scenario.



\bibliography{emnlp2020}
\bibliographystyle{acl_natbib}

\end{document}